\PassOptionsToPackage{table}{xcolor}
\documentclass[sigconf]{acmart}

\AtBeginDocument{%
  }

\copyrightyear{2025}
\acmYear{2025}
\setcopyright{acmlicensed}\acmConference[WWW Companion '25]{Companion Proceedings of the ACM Web Conference 2025}{April 28-May 2, 2025}{Sydney, NSW, Australia}
\acmBooktitle{Companion Proceedings of the ACM Web Conference 2025 (WWW Companion '25), April 28-May 2, 2025, Sydney, NSW, Australia}
\acmDOI{10.1145/nnnnnnn.nnnnnnn}
\acmISBN{979-x-xxxx-xxxx-x/xxxx/xx}

\usepackage{fontawesome5}
\usepackage{multirow}
\usepackage{bbding}
\usepackage{ulem}
\usepackage{amsmath}
\usepackage{amsfonts}
\usepackage{multirow}
\usepackage{booktabs}
\usepackage{dsfont}
\usepackage{algorithm}
\usepackage{algorithmic}
\definecolor{lightgreen}{RGB}{224,242,234} 
\definecolor{mygray}{rgb}{0.86, 0.86, 0.86}
\begin{document}

\title{\textit{One Size doesn't Fit All}: A Personalized Conversational Tutoring Agent for Mathematics Instruction}
\author{Ben Liu}
\authornote{Equal contribution.}
\affiliation{%
  \department{School of Computer Science,}
  \institution{Wuhan University,}
  \city{Wuhan}
  \country{China}
}
\email{liuben123@whu.edu.cn}

\author{Jihai Zhang}
\authornotemark[1]
\affiliation{%
  \department{DAMO Academy,}
  \institution{Alibaba Group,}
  \city{Hangzhou}
  \country{China}
}
\email{jihai.zjh@alibaba-inc.com}

\author{Fangquan Lin}
\affiliation{%
  \department{DAMO Academy,}
  \institution{Alibaba Group,}
  \city{Hangzhou}
  \country{China}
}
\email{fangquan.linfq@alibaba-inc.com}

\author{Xu Jia}
\affiliation{%
  \department{College of Computer and Cyber Security,}
  \institution{Hebei Normal University,}
  \city{Shijiazhuang}
  \country{China}
}
\email{jiaxu@hebtu.edu.cn}

\author{Min Peng}
\authornote{Corresponding Author.}
\affiliation{%
  \department{School of Computer Science,}
  \institution{Wuhan University,}
  \city{Wuhan}
  \country{China}
}
\email{pengm@whu.edu.cn}

\renewcommand{\shortauthors}{Ben Liu et al.}

\begin{abstract}
Large language models (LLMs) have been increasingly employed in various intelligent educational systems, simulating human tutors to facilitate effective human-machine interaction. However, previous studies often overlook the significance of recognizing and adapting to individual learner characteristics. Such adaptation is crucial for enhancing student engagement and learning efficiency, particularly in mathematics instruction, where diverse learning styles require personalized strategies to promote comprehension and enthusiasm. In this paper, we propose a \textbf{P}erson\textbf{A}lized \textbf{C}onversational tutoring ag\textbf{E}nt (PACE) for mathematics instruction. PACE simulates students' learning styles based on the Felder and Silverman learning style model, aligning with each student's persona. In this way, our PACE can effectively assess the personality of students, allowing to develop individualized teaching strategies that resonate with their unique learning styles. To further enhance students' comprehension, PACE employs the Socratic teaching method to provide instant feedback and encourage deep thinking. By constructing personalized teaching data and training models, PACE demonstrates the ability to identify and adapt to the unique needs of each student, significantly improving the overall learning experience and outcomes. Moreover, we establish multi-aspect evaluation criteria and conduct extensive analysis to assess the performance of personalized teaching. Experimental results demonstrate the superiority of our model in personalizing the educational experience and motivating students compared to existing methods.
\end{abstract}





\maketitle

\section{Introduction}
Intelligent Tutoring Systems (ITSs) are essential tools in education practice, providing immediate instruction and feedback to learners~\cite{ITS0,ITS1}. In recent years, conversational ITSs have attracted significant attention~\cite{conv_its,conv_its2,conv_its3,math} due to their ability to engage students through natural language interactions, thereby facilitating students in problem-solving by providing hints in text form. With large language models (LLMs) demonstrating impressive capabilities in simulating human behavior~\cite{char,oscars}, extensive research has been conducted to utilize LLMs as tutoring agents across various domains, including science~\cite{sci,boo2dial}, language learning~\cite{language}, and social skills coaching~\cite{social}. These LLM-driven conversational ITSs hold promising potential to transform students' learning experiences and significantly improve both engagement and knowledge acquisition.

\begin{figure}[ht]
    \centering
    \includegraphics[width=\linewidth]{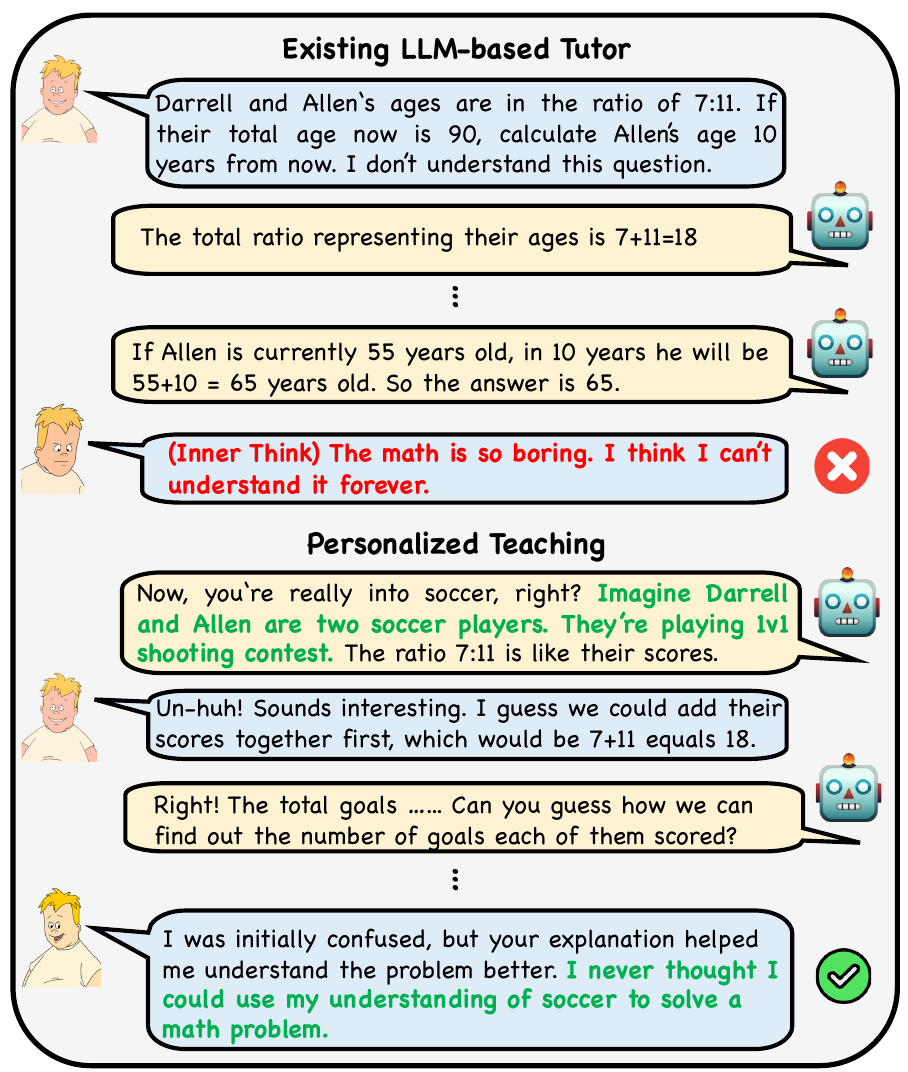}
    \caption{Comparison of personalized teaching with existing systems.}
    \label{intro}
\end{figure}

However, existing LLMs-driven ITSs rely on predetermined scaffolding strategies developed by human tutors~\cite{educhat,socratic}, which restrict their ability to address individual student needs and deliver tailored step-by-step guidance. The interests, personality traits, and experiences of students result in various learning styles, directly affecting their capacity to comprehend and absorb knowledge. This limitation is particularly obvious in mathematics instruction, where the subtleties of individual understanding and problem-solving approaches are essential for effective teaching. As shown in Figure~\ref{intro}, when mathematical concepts are connected to students' interests and illustrated with real-life examples or relatable scenarios, it can significantly boost their engagement and enthusiasm for mathematics. In contrast, traditional scaffolding strategies fail to personalize the learning experience, leading to a one-size-fits-all approach.

In this paper, we aim to develop a personalized LLM-driven tutor for mathematics instruction that not only guides students toward a deeper understanding but also aligns with their interests, personality traits, and experiences. Besides, we seek to dynamically adapt teaching strategies and content to varied dialogue contexts, ensuring that each student receives targeted support tailored to their individual needs. However, achieving this goal presents several significant challenges that require careful consideration: (1) A fine-grained approach is necessary to assess how specific personas influence teaching effectiveness, given their complexity. (2) Implementing tailored strategies that align with students' learning styles is essential for fostering engagement and critical thinking. (3) There is no established dataset for personalized mathematics teaching, and constructing such a dataset manually would be labor-intensive and costly, considering the diversity of personalities.

To tackle these challenges, we propose a \textbf{P}erson\textbf{A}lized \textbf{C}onvers- ational tutoring ag\textbf{E}nt (PACE), which generates responses tailored to the individual learning styles of students based on their personas. Instead of directly integrating students' personality traits into teaching strategies, PACE simulates students' learning styles based on their personas, employing the Felder and Silverman learning style model~\cite{felder}. This approach allows our model to generalize across diverse learners. Subsequently, PACE conceptualizes personalized teaching strategies that align with students' learning styles prior to teaching. Finally, PACE adopts the Socratic teaching method~\cite{socratic_teach} to encourage students to think critically, reflect, and explore concepts in depth rather than simply providing answers. Additionally, we collect representative student personas from the school television series \textit{Recess}, which encompass a diverse range of genders, interests, and levels of knowledge. Grounded in the teaching process outlined in PACE, we leverage the role-playing capabilities of GPT-4~\cite{gpt4} to simulate both teacher and student roles. This enables us to automatically synthesize personalized dialogues using an LLM-to-LLM interaction framework tailored to these persona profiles. Our contributions can be summarized as follows:
\begin{itemize}
    \item We introduce an LLM-driven personalized tutoring model, PACE, that effectively integrates students' personalities into teaching strategies, providing customized instruction for mathematics. Our model fosters critical thinking, comprehension, and reasoning, ultimately enhancing the overall learning experience and outcomes.
    \item We utilize LLMs to mimic tutor-student interactions grounded in our proposed teaching process and introduce an LLM-synthesized personalized teaching dataset.
    \item We establish multi-aspect evaluation criteria and implement comprehensive evaluation to assess the effectiveness of our model. Extensive experimental results demonstrate the superiority of our approach in personalizing the educational experience and motivating students compared to existing methods.
\end{itemize}

\section{Related Work}
\subsection{Intelligent Tutoring Systems}
Research has demonstrated the effectiveness of Intelligent Tutoring Systems (ITS) in improving student engagement and learning efficacy. Notably, \cite{bloom} emphasizes that ITS provide personalized instruction can lead to substantial improvements in student performance. This finding is supported by contemporary studies, such as \cite{stern}, which confirm that ITS can significantly enhance learning outcomes when compared to traditional classroom settings. A meta-analysis by \cite{zhuang} further highlights the effectiveness of ITS in education, showing considerable gains in students' problem-solving skills and knowledge retention.

Existing works design rule-based systems with human-crafted domain knowledge~\cite{rule}, or data-driven approaches~\cite{domain,Miao} that perform supervised learning on a certain amount of human annotation, enabling naturalistic exchanges between students and the system. This boosts student motivation and self-regulated learning, contributing to a more inclusive and engaging educational environment~\cite{student_model}.







\subsection{AI in Education}
Artificial Intelligence (AI) is transforming the educational landscape by enhancing learning experiences and personalizing educational journeys. AI technologies, such as machine learning, natural language processing, and data analytics, enable the development of adaptive learning systems, intelligent tutoring, and automated administrative tasks~\cite{Mzwri}. These innovations not only aid educators in identifying student needs but also create tailored learning paths that can improve engagement and outcomes.

Recent large language models show strong potential to build dialogue tutors with less data supervision and higher confidence~\cite{related_work, conv_its3, mathdial}. ITS can be further improved by integrating LLMs with pedagogical and learning science principles~\cite{educhat,boo2dial,class_agent}. Additionally, recent works~\cite{chi, personalize} demonstrate the potential of LLMs for individual student modeling.

\section{Methods}
In this section, we first present the framework of our PACE. We then introduce the dialogue synthesis process for personalized mathematical teaching. Finally, we enhance PACE's teaching capabilities by fine-tuning the LLM on curated dialogue data.

\begin{figure*}[t]
    \centering
    \includegraphics[width=\linewidth]{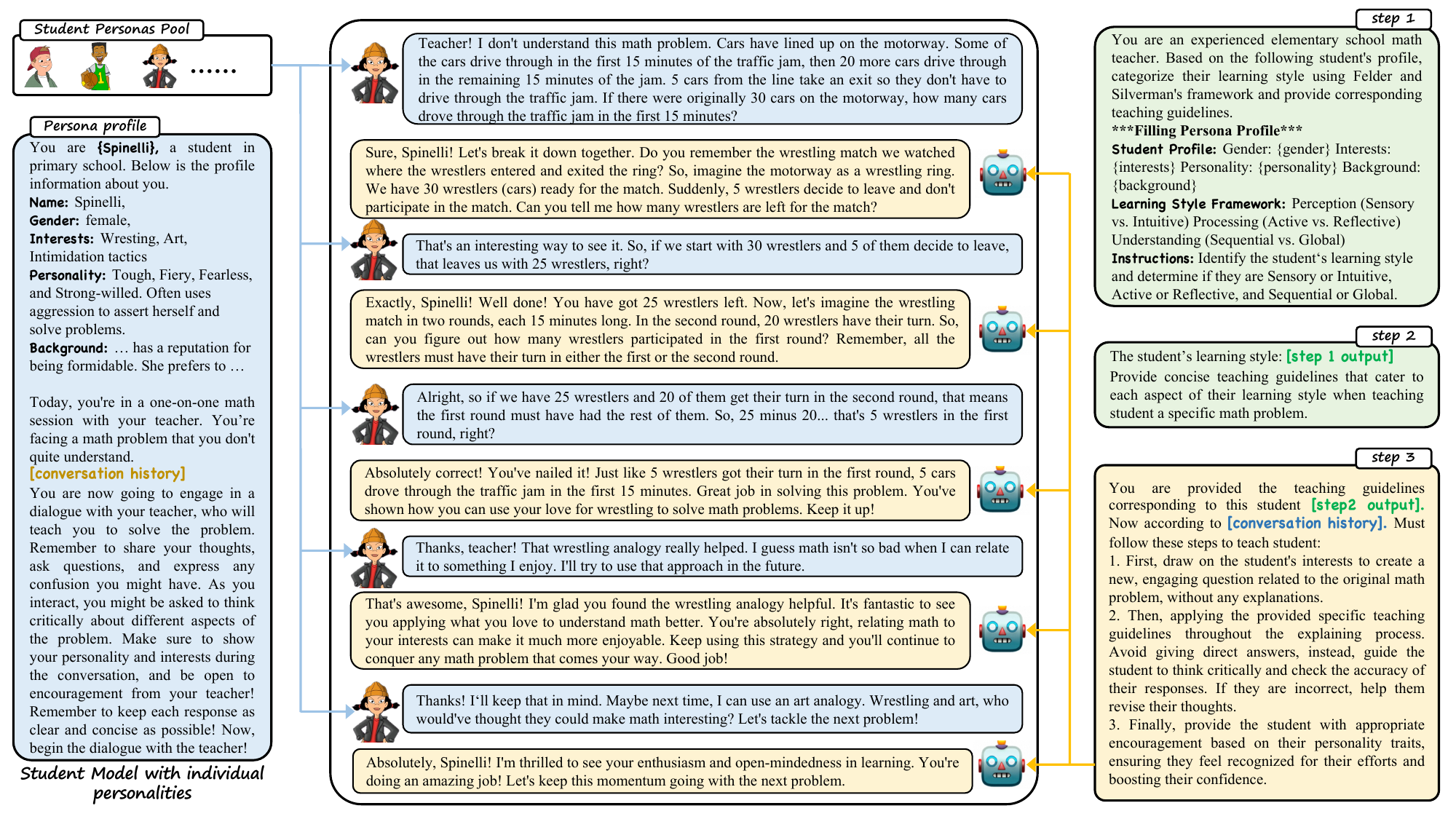}
    \caption{An example of how PACE communicates with a student and corresponding prompts both sides to generate expected dialogues as training data.}
    \label{framework}
\end{figure*}

\subsection{Personality-aware Tutoring Agent}
The framework of PACE, depicted on the right side of Figure~\ref{framework}, comprises three key stages: (1) simulating the learning style of each student based on their persona, (2) conceptualizing individualized teaching strategies that align with learning styles, and (3) guiding students toward profound thinking and stimulating their interest in mathematics through Socratic-style conversations. This streamlined process enhances personalized teaching and promotes a more engaging educational experience.

\subsubsection{Simulating Learning Styles} Personality is one of the most influential factors in education research~\cite{person2}, which significantly influences student engagement and knowledge acquisition. Students with different personas often display diverse learning styles, affecting how they absorb, process, comprehend, and retain information~\cite{felder,influ}. Therefore, we assess students' personalities by simulating their learning styles, enabling us to tailor teaching strategies effectively. 

We adopt the Felder and Silverman learning style model~\cite{felder}, which consists of eight main learning styles: Perception (\textit{sensory vs. intuitive}), Processing (\textit{active vs. reflective}), Understanding (\textit{sequential vs. global}), and Input (\textit{visual vs. auditory}). Since our agent relies on a dialogue-based interface, the Input learning style is constrained. Therefore, we focus on three distinct modes: Perception, Processing, and Understanding. Specifically, our PACE simulates students' learning styles based on their personalities, including gender, interests, characteristics, and experiences. The example in Figure~\ref{framework}, highlighted in green, we prompt the LLM to develop a nuanced understanding of each student's learning preferences, guided by the Felder and Silverman model. In this way, our PACE can effectively assess the personality of students, allowing us to develop individualized teaching strategies that resonate with their unique learning styles in subsequent sections.

\subsubsection{Conceptualizing Personalized Teaching Strategies} In this stage, we focus on developing tailored teaching strategies that align with the simulated learning styles of each student. By incorporating insights drawn from the Felder and Silverman model, our PACE aims to create engaging and effective mathematics instructions. For example, for students identified as sensory learners, we might implement strategies using physical objects or real-world examples to illustrate the problem. This hands-on approach helps sensory learners connect abstract concepts to tangible experiences, enhancing their understanding. Conversely, for intuitive learners, strategies include creating opportunities to discover patterns and relationships within mathematical concepts. This might involve activities that promote exploration and discovery, enabling these students to grasp overarching ideas and theoretical frameworks. Through this conceptualizing process, we integrate these personalized teaching guidelines into system prompts, adapting to each student's unique learning style. 

\subsubsection{Socratic-style Conversations} \label{socratic} After conceptualizing individualized teaching strategies, we adopt Socratic teaching method~\cite{socratic_teach}, encouraging students to think, reflect, and explore concepts deeply by asking thought-provoking questions rather than directly providing answers. Specifically, upon presenting a question, we first rephrase it based on the previously developed teaching strategies to make it more accessible for students. Subsequently, we propose prompt questions to help students navigate complex mathematical problems. Next, we evaluate their understanding by analyzing their responses and previous conversation records. If any misunderstandings arise, we address and correct them promptly. Finally, we repeat this process, allowing students to engage in cycles of reflection, correction, and prompting questions. This method not only enhances students' comprehension and critical thinking skills but also boosts their confidence and interest in mathematics.

\subsection{Personalized Teaching Conversation Construction} \label{data_construction}
Given the lack of personalized teaching datasets, we utilize the character-driven simulation capabilities of LLMs~\cite{oscars} to construct a multi-turn dialogue dataset with high quality. This process mainly involves three procedures: raw data collection, dialogue synthesis via LLMs, and human annotation.

\subsubsection{Raw Data Collection} The raw data is collected from an existing dataset, GSM8K~\cite{gms8k}. The math questions in GSM8K are high-quality grade school math problems created by human problem writers. These problems involve between 2 and 8 steps to solve and come with high-quality annotations, making them suitable for guiding students toward deeper thinking. Moreover, to build a simulation of students' personality traits, we need to collect different student characters, which span different genders, interests, and levels of knowledge. Thus, these characters should be quite representative for evaluating personalized teaching. For simplicity but without loss of generality, we construct six character profiles that exhibit distinctive traits. These characters are inspired by the protagonists of the school television series \textit{Recess}, all of whom are around ten years old and possess different personalities, hobbies, and experiences, making them well-suited for character modeling.

\subsubsection{Dialogue Synthesis via LLMs} We propose a method to simulate teacher and student personas by prompting LLMs for dialogue synthesis. Specifically, we utilize the GPT-4 (GPT-4-Turbo-8k) model to serve as the virtual teacher and student, respectively. For each math problem, a student profile is selected from the six characters to act as the system prompt for the student agent. The teacher agent, in turn, is tasked with simulating the student's learning styles based on the chosen profile and developing tailored teaching strategies during the conceptualizing process. This teaching strategy, combined with a Socratic teaching prompt, serves as the system prompt for the teacher agent. Consequently, a multi-turn dialogue is conducted between the teacher and student agents, fostering an interactive educational experience. 


\subsubsection{Human Annotation} To further enhance data quality and mitigate the impacts of randomness in LLMs, we manually clean and re-annotate the conversations. First, we retain only those dialogues that exceed five turns (where a turn consists of two utterances), filtering out shorter dialogues. Additionally, we review the final utterance of the student in each dialogue to eliminate samples in which the student’s problem remains unresolved. Subsequently, our domain-expert co-authors evaluate the coherence and quality of the dialogues, removing any problematic instances.

\subsection{PACE Training}
Personalized conversational tutoring can be viewed as an interactive task, where the objective of the agent model is to generate responses based on the student's persona profile and the conversation history. Formally, given the task instruction $\tau$ and the student's profile $p$, our language agent PACE with parameters $\theta$ serves as the policy model $\pi_{\theta}$, responsible for generating a tailored response $r_{t+1}$ based on historical conversation $h_t$ at the $t$-th turn:
\begin{equation}
    \begin{split}
    &r_{t+1} \sim \pi_\theta(\cdot \vert h_t, p), \\
        &h_t = (\tau, u_0, r_0, u_1, r_1, \ldots, u_t, r_t),
    \end{split}
\end{equation}
where $u$ and $r$ are the utterance and response from the student and agent, respectively.

To foster an engaging learning experience that not only aligns with students' personality traits but also resonates with their interests, we steer the PACE to self-synthesize the learning styles of students based on their personality profiles. Subsequently, we prompt the PACE to generate teaching strategies $s$ according to the simulated learning styles $l$. This process can be represented as:
\begin{equation}
    s \sim \pi_\theta (\cdot | \rho_{strategy}, l)\pi_\theta (l | \rho_{style}, \tau, p),
\end{equation}
where $\rho_{style}$ stands for the prompt to instruct the learning style simulation based on the Felder and Silverman learning style model, and $\rho_{strategy}$ denotes the prompt to instruct the personalized teaching strategies summarization. 

Ultimately, the whole conversation trajectory $c$ concludes when the student's problem is solved or exceeds the maximum dialogue turns. The entire trajectory with turn size $n$ can be modeled as follows:
\begin{equation}
    \pi_{\theta}(c| \tau) = \prod \limits_{t=0}^n \pi_\theta (r_{t + 1} | h_t, s)\pi_\theta(r_0|u_0, s, \tau).
\end{equation}

Therefore, given the expert conversation trajectories dataset $\mathcal{D}=\{(\tau, s, c)^{(i)}\}^{|\mathcal{D}|}_{i=1}$, we train our PACE to follow the personalized teaching strategies $s$ to generate tailored responses. Under an auto-regressive manner, the loss of the PACE can be formulated as:
\begin{equation}
    \mathcal{L}_{PACE}(\pi_\theta)=-\mathbb{E}_{c\sim \mathcal{D}}[\pi_\theta(c|\tau, s)].
\end{equation}
Suppose $\mathcal{X}=(x_0, x_1, \ldots, x_{|\mathcal{X}|-1})$ is the token sequence of the conversation trajectory $c$, we only compute the loss on tokens that belong to teachers:
\begin{equation}
    \pi_\theta(c|\tau, s) = -\sum_{j=1}^{|\mathcal{X}|-1}(\mathds{1}(x_j \in r) \times \log \pi_\theta(x_j|\tau, s, x_{< j})),
\end{equation}
where $\mathds{1}(x_j \in r)$ is the indicator function to mask tokens unrelated to teacher responses.

\section{Multi-aspect Evaluation Criteria}
\begin{figure}[t]
    \centering
    \includegraphics[width=0.8\linewidth]{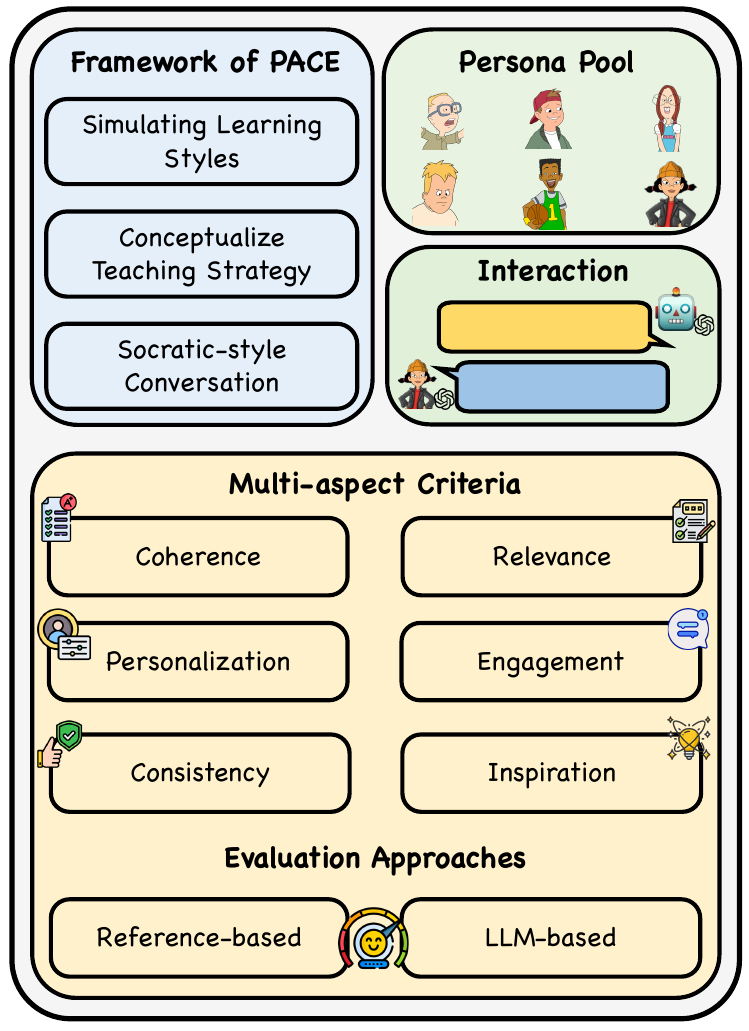}
    \caption{The overall design framework of our PACE.}
    \label{overall}
\end{figure}

In this section, we outline the evaluation criteria that the responses of the PACE should meet and identify potential factors within the dialogue that may reflect the quality of teaching. To provide a comprehensive assessment, we employ a dual evaluation approach consisting of reference-based and LLM-based methodologies. The overall framework of our PACE can refer to Figure~\ref{overall}.

\subsection{Reference-based Evaluation}
Reference-based evaluation assesses the model's responses against provided reference output (expert response or ground truth) to determine whether they meet the established criteria. This approach is based on the assumption that the greater the similarity between the model's outputs and the references, the more consistent they are with the desired qualities. To this end, we adopt several metrics, including \textit{BLEU}~\cite{bleu} series, \textit{ROUGE}~\cite{rouge} series, \textit{METEOR}~\cite{meteor}, and \textit{BERTScore}~\cite{bertscore}, which are standard evaluation metrics commonly used in traditional text generation tasks. BLEU-$n$ assesses the $n$-gram precision of the generated text, while ROUGE-$n$ measures $n$-gram recall, assessing the degree of similarity between the model's output and the reference output. METEOR is a multi-factorial evaluation method that assesses the quality of the generated text in terms of precision, recall and matching success. BERTScore leverages contextual embeddings from the BERT~\cite{bert} to evaluate the similarity between generated text and reference output,  allowing for a more accurate and unbiased assessment compared to BLEU and ROUGE.

\begin{table*}[t]
    \centering
    \begin{tabular}{lcccccccccc}\toprule
    \multirow{2}{*}{\textbf{Methods}} & \multicolumn{10}{c}{\textbf{Referenced-based Evaluation}}\\\cmidrule(lr){2-11}
     & BL-1 & BL-2 & BL-3 & BL-4 & METEOR &  RG-1 & RG-2 & RG-L & BERTScore & \textbf{Avg.} \\\midrule
     \textit{LLaMA2-7B-chat} & 25.68 & 14.22 & 8.97 & 5.79 & 32.62 & 33.99 & 11.04 & 20.91 & 63.97 & 24.13 \\
     \textit{LLaMA2-70B-chat} & 27.59 & 16.36 & 10.87 & 7.20 & 34.47 & 34.54 & 11.86 & 21.67 & 66.57 & 25.68 \\
     \textit{Mistral-7B-Instruct} & 23.80 & 12.64 & 7.85 & 5.08 & 27.56 & 29.30 & 9.16 & 19.01 & 62.55 & 21.88\\
     \textit{Mistral-8$\times$7B-Instruct} & 24.79 & 14.93 & 9.91 & 6.52 & 35.44 & 33.50 & 11.90 & 21.09 & 66.79 & 24.99\\
     \textit{Qwen2-7B-Instruct} & 26.51 & 15.90 & 10.55 & 6.96 & 34.62 & 34.42 & 12.43 & 21.95 & 66.54 & 25.54 \\
     \textit{Qwen2-72B-Instruct} & 28.49 & 17.83 & 12.26 & 8.34 & \textbf{38.62} & 37.33 & 14.53 & 23.82 & 68.05 & 27.70\\\midrule
     \textit{EduChat} & 21.38 & 11.33 & 7.08 & 4.58 & 20.77 & 27.94 & 8.65 & 18.92 & 59.11 & 19.97 \\\midrule
     \rowcolor{lightgreen}
     \textit{PACE} & \textbf{39.92} & \textbf{27.31} & \textbf{20.25} & \textbf{14.98} & \underline{37.58} & \textbf{45.65} & \textbf{20.11} & \textbf{33.20} & \textbf{71.99}  & \textbf{34.55}\\
     \rowcolor{mygray}
     \textit{w/o student simulation} & 38.24 & 25.31 & 18.31 & 13.13 & 35.60 & 43.80 & 17.72 & 31.18 & 70.59 & 32.65\\
     \rowcolor{mygray}
     \textit{w/o socratic teaching} & \underline{38.75} & \underline{25.99} & \underline{18.89} & \underline{13.61} & 36.38 & \underline{44.47} & \underline{18.63} & \underline{31.90} & \underline{71.31} & \underline{33.33}\\\bottomrule
    \end{tabular}
    \caption{Performances of different methods. BL-n, RG-n, and RG-L denotes scores ($\%$) of BLEU-n, ROUGE-n, and ROUGE-L.}
    \label{reference_res}
\end{table*}

\subsection{LLM-based Evaluation}

Beyond conventional objective metrics, evaluating personalized conversational tutoring models requires a nuanced framework that assesses their ability to deliver customized teaching experiences across multiple dimensions.  Given the cost and bias inherent in human evaluations, we utilize LLMs as evaluators, proven effective in previous research~\cite{llm_eval}. To improve evaluation accuracy, our domain expert co-authors design and review 30 random dialogues to establish scoring examples, which we use to guide GPT-4 in performing evaluations. We assess the responses of our PACE based on six criteria:

\begin{itemize}
    \item \textit{Coherence.} It evaluates the degree to which the response is logically consistent with the ongoing conversation. High coherence indicates that the dialogue flows naturally, with each response being contextually appropriate and maintaining continuity throughout the interaction.
    
    \item \textit{Relevance.} This criterion ensures that the information provided is pertinent to the context of the conversation and goes beyond general knowledge to meet individual student inquiries.
    
    \item \textit{Personalization.} This dimension emphasizes the model's ability to adapt its interactions based on preferred learning styles and the specific characteristics of the student.

    \item \textit{Engagement.}  This metric evaluates whether the interaction fosters a positive learning atmosphere and encourages active participation from the students.
    
    \item \textit{Consistency.} Evaluate if the model's statements match or contradict the student's learning styles and prior interactions. This criterion ensures that the model maintains a stable approach that resonates with students throughout the teaching process.
    
    \item \textit{Inspiration.} Inspiration ensures the responses motivate students, sparking curiosity and encouraging further exploration.

\end{itemize}


\section{Experiments}
\subsection{Experimental Settings}
\subsubsection{Dataset} As outlined in Method section,  we collect student personas from the school television series \textit{Recess}, which demonstrate a variety of genders, hobbies, personality traits and experiences. After selecting these personas, we prompt GPT-4 with the temperature of 0.7 and the top-p of 0.95 to generate dialogue data, simulating interactions between students and the teacher. The mathematical questions are sourced from GSM8K~\cite{gms8k}, which includes 8,792 questions requiring between 2 to 8 steps to solve. After human annotation, our constructed dataset comprises 1,410 dialogues, focusing on 6 representative personas. The dataset is divided into 1,200/60/150 for training, validation, and testing. The detailed statistic is shown in Table~\ref{dataset}.
\begin{table}[h]
    \centering
   \begin{tabular}{l|l}\toprule
    \textbf{Dataset Summary} & \textbf{Counts} \\\midrule
    Dialogues & 1,410 \\
    Turns & 14,100 \\
    Avg. Turns per dialogue & 10 \\
    Avg. Words per utterance (Tutor) & 54.56 \\
    Avg. Words per utterance (Student) & 44.67 \\\bottomrule
    \end{tabular}
    \caption{Constructed tutoring dataset summary.}
    \label{dataset}
\end{table}

\subsubsection{Baselines} We conduct experiments with several representative LLMs: LLaMA2-chat (7B and 70B)~\cite{llama}, Mistral (7B and 8$\times$7B)~\cite{mistral,moe}, and Qwen2-Instruct (7B and 72B)~\cite{qwen}. Following methodologies proposed in previous research~\cite{chi}, we employ personality-aware instructions as prompts to encourage each model to act as a mathematics tutor. Additionally, we compare PACE with EduChat~\cite{educhat}, an educational LLM built on the LLaMA2-7B architecture, which has been augmented with a corpus of educational conversation data obtained through supervised fine-tuning.

\subsubsection{Implementation Details}We employ the LLaMA2-7B-chat as the backbone of PACE. And we train the model using the LoRA~\cite{lora} approach with $r=32$ and $\alpha=32$. AdamW~\cite{adam} optimizer with learning rate of 3e-4 and warm ratio of $0.01$ is utilized. All experiments are carried out on a system equipped with two NVDIA A800 GPUs.

\subsection{Results of Reference-based Evaluation} 
Table~\ref{reference_res} displays the results of our experiments. Overall, we can observe that PACE achieves consistent and significant improvement across all metrics, which demonstrates the effectiveness of our proposed PACE. Compared to the outcomes of representative LLMs that utilize students' personalities as prompts, our performance significantly surpasses theirs, showing an average improvement ranging from $24.7\%$ to $43.2\%$. This demonstrates the superiority of our approach, which models students' learning styles and generates tailored teaching strategies. Additionally, our experimental results reveal that Educhat, despite being trained on a large-scale dataset of teaching dialogues, underperforms in personalized teaching scenarios. This shortcoming arises because Educhat focuses more on problem-solving than on considering the unique characteristics of students to engage them in the learning process effectively.

To further investigate, we conduct an ablation study to assess the impact of key components in PACE quantitatively. This involves removing the simulation of learning styles and the conceptualization of tailored teaching strategies (\textit{w/o student simulation}), as well as the Socratic teaching method (\textit{w/o socratic teaching}). The results indicate that all implemented modules contribute effectively to training outcomes. Notably, the absence of the student learning style simulation phase significantly affects the final results, highlighting the importance of considering students' learning styles in the educational process.

\subsection{Results of GPT4-based Evaluation}

\begin{figure}[ht]
    \centering
    \includegraphics[width=0.8\linewidth]{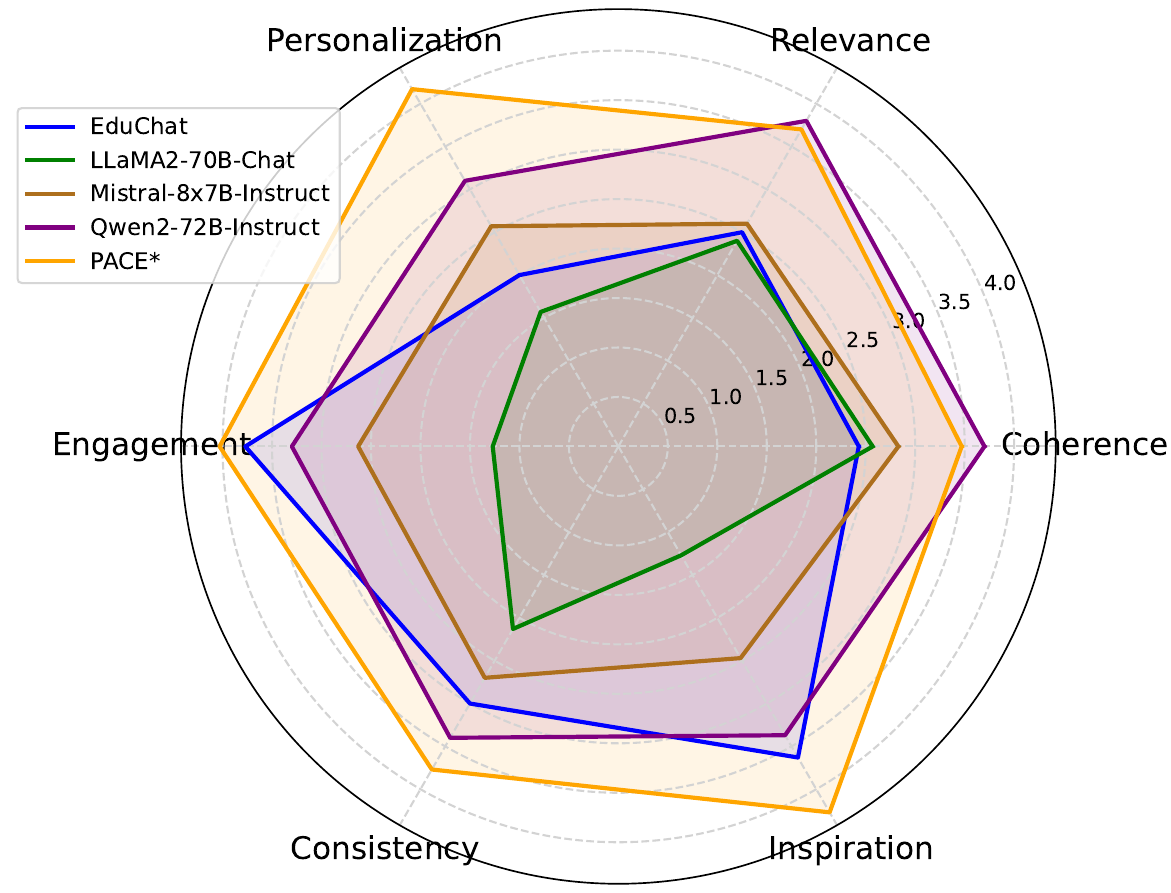}
    \caption{Multi-aspect assessment comparisons using GPT-4. Higher values indicate better performance.}
    \label{llm-based}
\end{figure}
Single reference-based automatic metrics do not always accurately reflect the real quality of the generated responses. Therefore, we conduct GPT-4-based evaluation, focusing on six aspects outlined in the Multi-aspect Evaluation Criteria Section. To mitigate potential consistency issues when assigning scores for all responses individually, we adopt an alternative approach. We instruct GPT-4 to \textbf{rank} the generated responses based on the conversation history and students' persona. This ranking process enables a more effective evaluation of the performance across different models. Additionally, we include five ranking examples, curated by our domain-expert co-authors, as demonstrations for each assessment criterion. Detailed prompts can refer to Table~\ref{evaluation prompt}.
\begin{table}[h]
    \centering
    \resizebox{\linewidth}{!}{
    \begin{tabular}{lp{0.8\linewidth}}
    \toprule
    \textbf{Criteria} & \textbf{Simplified Version of Prompt} \\
    \midrule
    Coherence & ...Please rank the responses of each model based on coherence. When evaluating, consider the logical consistency of the responses with each other, the clarity and flow of the dialogue, and whether there are any abrupt topic changes or gaps in the conversation.  \textit{Model Dialogues: Model 1, Model2, ..., Model 5.} Please provide only the ranking from the lowest to highest. \\\midrule
    Relevance & ...Focus on how well each response addresses the specific questions posed by the student, the pertinence of the information provided to the context of the dialogue, and whether the responses go beyond general knowledge to meet individual student inquiries... \\\midrule
    Personalization & ...Pay attention to how well each model adapts its responses to the student's personality traits, the consideration of the student's learning styles in the responses, and any evidence of understanding the unique characteristics of the student...\\\midrule
    Engagement & ...Consider how effectively the responses encourage active participation from the student, the creation of a positive and stimulating atmosphere during the conversation, and any strategies used to maintain or boost the student's interest in the topic...\\\midrule
    Consistency & ...Focus on whether the responses align with the student's previously expressed preferences and learning styles, the absence of contradictions within the responses or with previous interactions, and the maintenance of a coherent and stable approach throughout the dialogue... \\\midrule
    Inspiration & ...Pay attention to whether the responses motivate the student and spark curiosity, the sense of encouragement for further exploration and inquiry, and how well the model employs techniques that inspire critical thinking or enthusiasm for the subject... \\\bottomrule
    \end{tabular}
    }
    \caption{Prompt template for criteria that require the use of GPT-4 evaluation.}
    \label{evaluation prompt}
\end{table}

We compare our PACE model with the 70B series model due to their better performance in reference-based evaluation. As shown in Figure~\ref{llm-based}, PACE significantly outperforms the existing models across all dimensions, particularly in personalization, engagement, and inspiration, underscoring the superiority of our model. Notably, although EduChat underperforms in the reference-based assessment, it meets expectations in the LLM-based evaluation, highlighting the importance of LLM-based assessments. Overall, in both evaluation types, PACE demonstrates exceptional personalized teaching capabilities, crucial for enhancing student engagement and inspiration in mathematics education.

\subsection{Performance on Unseen Student Persona}
To further explore the benefits of simulating students' learning styles, we evaluate the performance of PACE across various unseen student personas. We construct a new set of student personas that significantly diverge from those in our previously established dataset. We then utilize GPT-4 to simulate these personas, re-collect a new set of mathematical questions, and engage our model, along with baselines, in dialogue with these unseen, GPT-4 simulated students. In this scenario, models are required to provide fine-grained assessments of students to improve the teaching experience.

\begin{figure}[h]
    \centering
    \includegraphics[width=\linewidth]{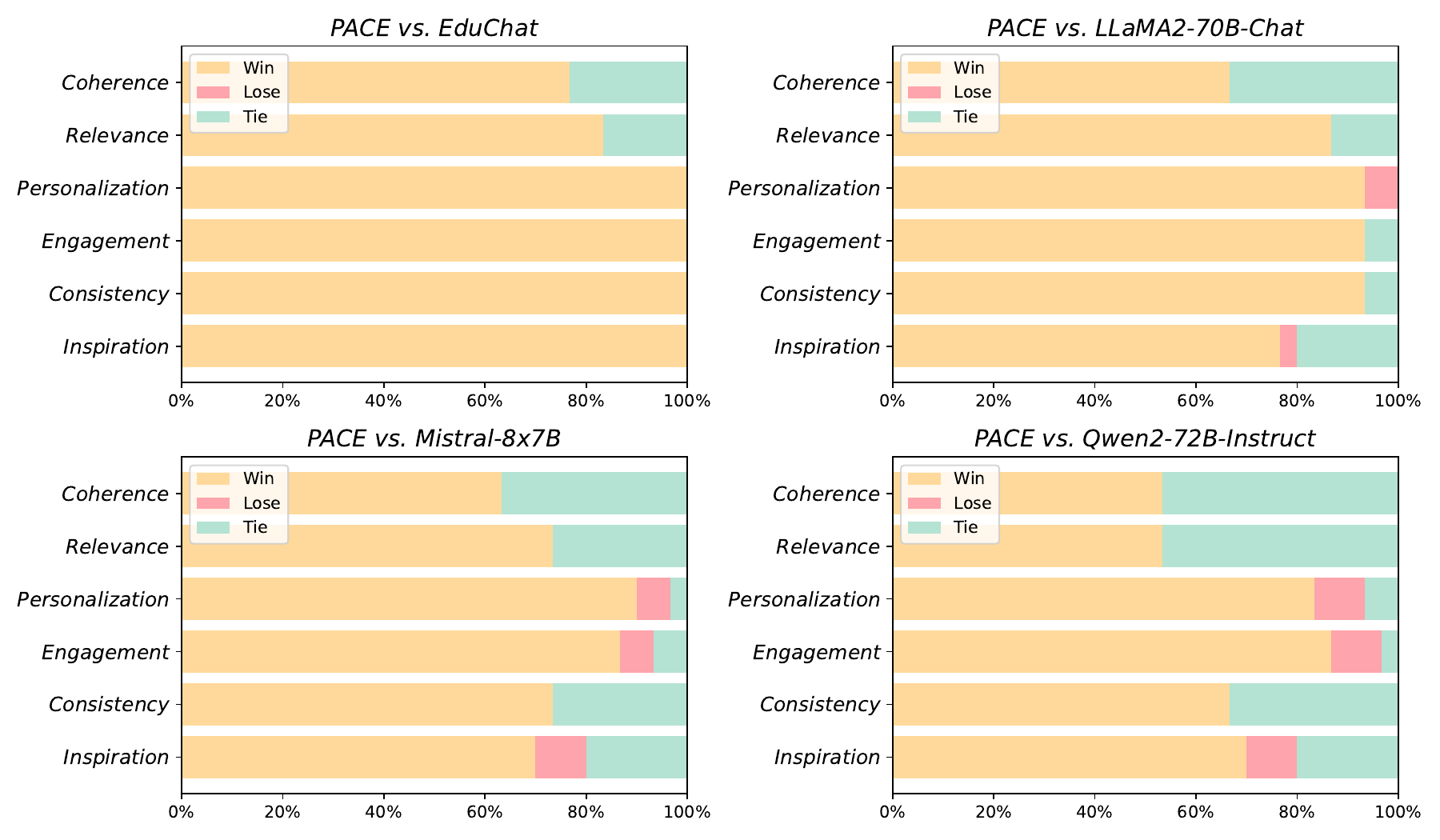}
    \caption{Evaluation of different models on unseen student personas across defined criteria.}
    \label{adv perf}
\end{figure}

To effectively evaluate the performance of each model, we employ adversarial evaluation techniques~\cite{gpt4}. For each response pair generated by the two models, we utilize GPT-4 to determine which response better meets the specified assessment criteria, categorizing the results as win, lose, or tie. The experimental results, illustrated in Figure~\ref{adv perf}, demonstrate that PACE exhibits strong generalization capabilities compared to existing models. This can be attributed to our implementation of Felder and Silverman learning style model, which enables PACE to simulate students' learning styles and generate tailored teaching strategies, rather than relying on a static teaching approach. The experimental results indicate that the proposed framework facilitates personalized teaching rather than merely memorizing predefined strategies.

\section{Conclusion}
In this paper, we present PACE, a novel LLM-based framework designed to enhance personalized mathematics instruction. By simulating diverse student personas based on the Felder and Silverman learning style model, PACE tailors instructional strategies to meet individual learning preferences. The incorporation of the Socratic teaching method further promotes critical thinking and deeper engagement with mathematical concepts. Utilizing the role-simulation capabilities of LLMs, we construct a dataset for personalized instruction and propose comprehensive evaluation metrics. Experimental results demonstrate the effectiveness of our framework in delivering tailored educational experiences, leading to increased student engagement and understanding. 

\begin{acks}
We would like to thank all the anonymous reviewers and area chairs for their comments. This work is supported by DAMO Academy through DAMO Academy Research Intern Program. This work is supported by the National Natural Science Foundation of China (U23A20316) and funded by the Joint\&Laboratory on Credit Technology.
\end{acks}

\bibliographystyle{ACM-Reference-Format}
\bibliography{sample-base}


\end{document}